\documentclass{bmvc2k}

\usepackage{graphicx}
\usepackage{amsmath}
\usepackage{amssymb}
\usepackage{booktabs}

\DeclareMathOperator*{\argmax}{arg\,max}

\newcommand{\overbar}[1]{\mkern 1.5mu\overline{\mkern-1.5mu#1\mkern-1.5mu}\mkern 1.5mu}

\title{Localizing Actions from Video Labels\\and Pseudo-Annotations}

\addauthor{Pascal Mettes}{}{1}
\addauthor{Cees G.M. Snoek}{}{1}
\addauthor{Shih-Fu Chang}{}{2}

\addinstitution{
 University of Amsterdam\\
 Amsterdam, NL
}
\addinstitution{
 Columbia University\\
 New York, USA
}

\runninghead{Mettes, Snoek, Chang}{Action Localization with Pseudo-Annotations}

\def\etal{\emph{et al}\bmvaOneDot}

\begin{document}

\maketitle

\begin{abstract}
The goal of this paper is to determine the spatio-temporal location of actions in video.
Where training from hard to obtain box annotations is the norm, we propose an intuitive and effective algorithm that localizes actions from their class label only.
We are inspired by recent work showing that unsupervised action proposals selected with human point-supervision perform as well as using expensive box annotations.
Rather than asking users to provide point supervision, we propose fully automatic visual cues that replace manual point annotations.
We call the cues pseudo-annotations, introduce five of them, and propose a correlation metric for automatically selecting and combining them.
Thorough evaluation on challenging action localization datasets shows that we reach results comparable to results with full box supervision.
We also show that pseudo-annotations can be leveraged during testing to improve weakly- and strongly-supervised localizers.
\end{abstract}

\section{Introduction}
The goal of this paper is to determine the spatio-temporal location of actions such as \emph{Skateboarding} and \emph{Shaking hands} in video content. This challenging problem is typically solved by classifying sliding cuboids~\cite{yanke_iccv05,lan2011discriminative,TianPartCVPR2013} action proposals~\cite{jain2014action,chencorsoICCV2015actiondetectionMotionClustering,oneata2014spatio,soomroICCV2015actionLocContextWalk,sultani2016if}, or by linking detectors over time~\cite{kang2016object,yuCVPR2015fap,weinzaepfelICCV2015learningToTrack,saha2016deep}.
In all cases, precise box annotations for actions on training video are a prerequisite for localizing actions in test videos.
We challenge the need for spatio-temporal box annotations and propose an intuitive and effective algorithm that localizes actions in video from a video label only.

We are inspired by the recent work of Mettes \etal~\cite{mettes2016spot}. For their training they start from unsupervised action proposals~\cite{vangemert2015apt}, typically about 1,000 sequences of bounding boxes that are generated automatically for a video. Mettes \etal~\cite{mettes2016spot} show that using the best possible action proposal during training, rather than ground truth annotations, does not lead to a decrease in action localization accuracy. Encouraged by this observation, they introduce a variant of the Multiple Instance Learning algorithm~\cite{andrews2002support} able to mine proposals with a good spatio-temporal fit to actions of interest by letting humans annotate a limited amount of points on the action in relevant training frames. While surprisingly effective, their approach still demands human supervision beyond the action class label. In this paper, we also rely on unsupervised action proposals during training, but rather than selecting the best proposal using manual human point-supervision we prefer a completely automatic alternative.

\begin{figure}[t]
\centering
\includegraphics[width=0.95\textwidth]{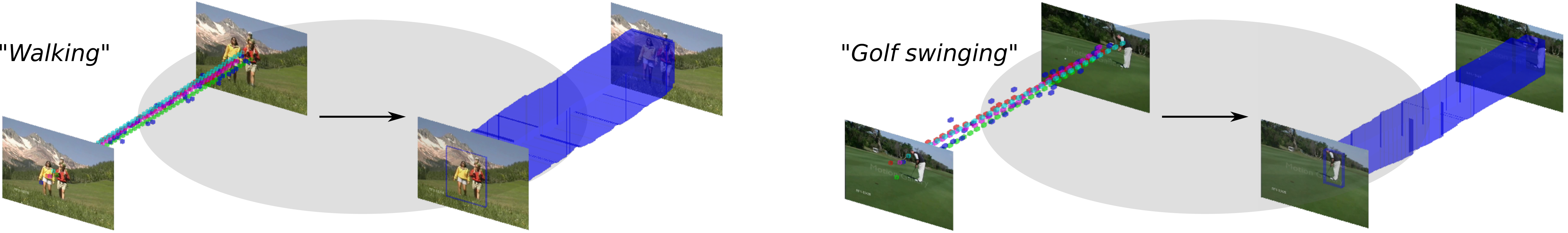}
\caption{\textbf{We introduce pseudo-annotations from visual cues,} indicated by different colored dots, that simulate supervision in videos. From the pseudo-annotations and action class labels, we automatically select action proposals (blue tube) for training action localizers.}
\label{fig:fig1}
\end{figure}

We introduce the notion of pseudo-annotations, see Figure~\ref{fig:fig1}, which we define as visual cues that replace point-supervision in video.
We investigate five of such pseudo-annotations by exploiting sources such as action proposals \cite{vangemert2015apt}, object proposals \cite{zitnick2014edge}, person detections \cite{yuCVPR2015fap}, motion \cite{jain2014action}, and center biases \cite{tseng2009quantifying} to discover which cues are most informative to point on the action locations.
The pseudo-annotations specify the likely location of an action in a video, resulting in the automatic selection of a desirable action proposal during Multiple Instance Learning optimization, where the information from pseudo-annotations is combined with action-specific video labels.
To automatically select and combine pseudo-annotations from different cues, we introduce a metric based on correlations between the pseudo-annotations.

Thorough evaluation on multiple action localization datasets shows that individually, each visual cue is informative for localizing actions.
Using our correlation metric for selecting and combining annotations, we reach results comparable to action localization from full box supervision with the same proposal and classification settings, while outperforming other weakly-supervised alternatives.
Furthermore, we demonstrate how pseudo-annotations can be leveraged during testing, to further improve any localization result, be it trained on pseudo-annotations or manually annotated boxes.

\section{Related work}

\indent
Yu and Yuan~\cite{yuCVPR2015fap} introduce supervised actor proposals for action localization. They rely on a person detector on successive frames and generate spatio-temporal proposals by assuring sufficient overlap and appearance consistency.
Gkioxari and Malik~\cite{gkioxari2015finding} replace the person detector by an action-specific detector using appearance and motion.
They link regions with strong overlap over time.
Weinzaepfel \etal~\cite{weinzaepfelICCV2015learningToTrack} follow the same scheme, but rather than linking detections they prefer tracking by detection (using boxes and class label) for further fine-tuning over time.
It is obvious that by adding more supervision to the action proposal generation, better localization can be achieved, especially with deep learning, see Saha \etal \cite{saha2016deep}.
Rather than using class-specific action detectors and box supervision, we prefer to localize an action in video from its class label only.

Jain \etal~\cite{jain2014action} introduce unsupervised action proposals that are likely to include the action, ideally achieving high recall with few proposals.
They start from super-voxels and group them based on color, texture, motion, size, fill cues, and independent motion.
Van Gemert \etal~\cite{vangemert2015apt} bypass the computationally expensive segmentation step of~\cite{jain2014action} by creating unsupervised proposals directly from dense trajectories~\cite{wang13} used to represent videos during classification. Chen and Corso~\cite{chencorsoICCV2015actiondetectionMotionClustering} also advocate clusters of dense trajectories for unsupervised action proposals. We also rely on unsupervised action proposals, but rather than selecting the best proposals using a classifier that learns from box annotations, we learn from a class label only.

Mettes \etal \cite{mettes2016spot} propose to train action localization classifiers using unsupervised proposals as positive examples rather than ground truth boxes. They introduce a Multiple Instance Learning (MIL) algorithm that mines proposals with a good spatio-temporal fit to actions by including point supervision. It extends the traditional MIL objective with a measure that takes into account the overlap between proposals and points. Their approach allows to localize actions in video from class labels and point annotations.
We also exploit a MIL optimization, but rather than relying on point-supervision, we prefer automated cues that do not require any action localization supervision.


\begin{figure*}[t]
\centering
\subfigure[]{\includegraphics[width=0.155\textwidth]{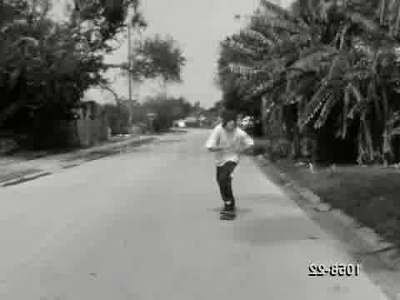}}
\hspace{0.1cm}
\subfigure[]{\includegraphics[width=0.155\textwidth]{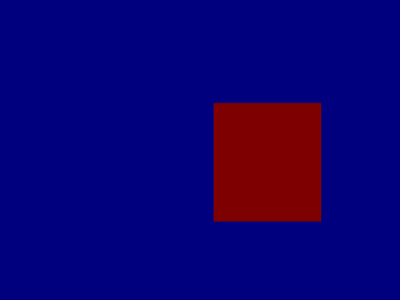}}
\subfigure[]{\includegraphics[width=0.155\textwidth]{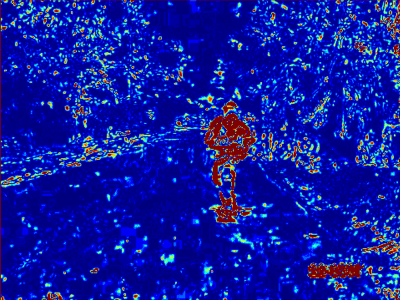}}
\subfigure[]{\includegraphics[width=0.155\textwidth]{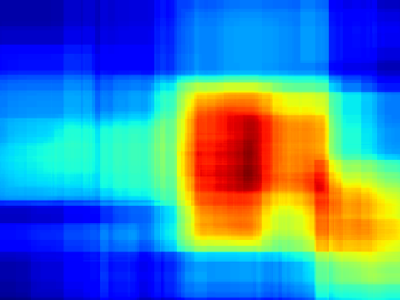}}
\subfigure[]{\includegraphics[width=0.155\textwidth]{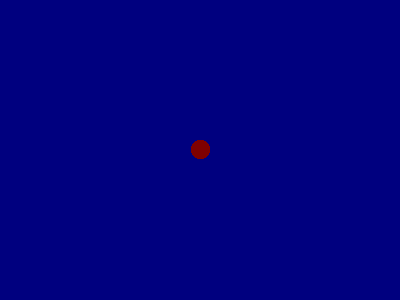}}
\subfigure[]{\includegraphics[width=0.155\textwidth]{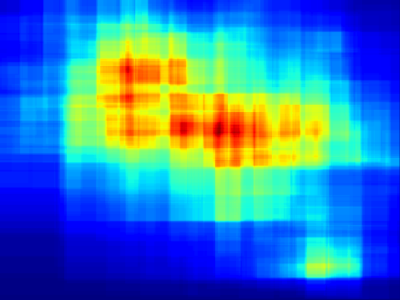}}
\caption{\textbf{Heatmaps of each of the five pseudo-annotations} for (a) an example frame from \emph{Skateboarding}.
From left to right: (b) person detection, (c) independent motion, (d) action proposals, (e) frame center, and (f) object proposals.}
\label{fig:cues}
\end{figure*}

\section{Action localization with Pseudo-annotations}
For training an action localizer, we are given a set of $N$ training videos $\{X_i,Y_i\}_{i=1}^{N}$, where $X_i \in \mathbb{R}^{|A_{i}| \times D}$ states the $|A_i|$ action proposals, each of feature dimension $D$, and $Y_i \in \{-1,+1\}$ indicates the video label, which is $+1$ if the action occurs anywhere in the video and $-1$ otherwise.
Each action proposal $A_i = \{A_{i}(t)\}_{t=1}^{T}$ is a tube consisting of $T$ bounding boxes.

Our goal is to train a classifier using a proposal with high spatio-temporal overlap with the action of interest for each video. We employ a Multiple Instance Learning perspective~\cite{andrews2002support,cinbis2014multi,mettes2016spot}. Each video is a bag and the proposals in each video are the instances. Using a max-margin objective, the Multiple Instance Learning optimization is given as:
\begin{equation}
\begin{split}
 & \min_{\mathbf{w},b,\xi} \frac{1}{2} ||\mathbf{w}||^{2} + \lambda \sum_{i} \xi_{i},\\
\text{s.t.} \quad & \forall_{i} : Y_{i} \cdot ( \mathbf{w} \cdot \argmax_{\mathbf{z} \in X_{i}} S(\mathbf{z} | \mathbf{w},b,P)) \geq 1 - \xi_{i}, \quad \forall_{i} : \xi_i \geq 0\\
\end{split}
\label{eq:milsvm}
\end{equation}
where $S(\mathbf{z} | \mathbf{w},b,P)$ specifies a selection function for proposal $\mathbf{z} \in X_i$, conditioned on both the classifier score ($\mathbf{w},b$) and (pseudo-)annotations $P$.

In this work, we only require video labels. Therefore, we are tasked with automatically discovering annotations $P$, dubbed pseudo-annotations.
They exploit sources such as action and object proposals, motion, humans, and center biases, see Figure~\ref{fig:cues}.
First, we outline how to obtain each individual pseudo-annotation, after which we show how to compute overlap scores from pseudo-annotations and how to combine the pseudo-annotations.

\subsection{Pseudo-annotations}

\indent \textbf{Pseudo-annotations from person detection.}
Actions are typically human-centered, so a robust detection of people in video frames provides information about the spatio-temporal location of actions. Here, we employ the Faster R-CNN network~\cite{ren2015faster}, using the person class after pre-training on MS-COCO~\cite{lin2014microsoft}.
After non-maximum suppression, the network yields roughly 50 box detections per frame, each with a confidence score. We select the bounding box in each frame with maximum confidence score as our pseudo-annotation.

\textbf{Pseudo-annotations from independent motion.}
The independent motion at each pixel of a frame $F$ provides information as to where the foreground action occurs in the frame. Here, we employ the interpretation of independent motion from Jain \etal~\cite{jain2014action}. Independent motion states the deviation from the global motion present in a frame. Let $\xi_{(x,y,F)} \in [0,1]$ denote the inverse of the residual in the global motion estimation at pixel $(x,y)$ in frame $F$. The higher the value of $\xi_{(x,y,F)}$, the less likely that the pixel contributes to the global motion. Then we compute a point-wise pseudo-annotation for frame $F$ as the center of mass over all the pixels in the frame, where the mass is given by their independent motion estimation:
\begin{equation}
p_{im}(F) = \frac{1}{\xi_{(F)}} \sum_{(x,y) \in F} \xi_{(x,y,F)} \cdot (x, y),
\end{equation}
where $\xi_{(F)}$ denotes the total independent motion in frame $F$.

\textbf{Pseudo-annotations from action proposals.}
We furthermore examine the action proposals themselves as a source of information for pseudo-annotations, using the unsupervised spatio-temporal proposals of~\cite{vangemert2015apt}. For a frame $F$ and action proposals $A^{*}$, we examine the spatial distribution of the proposal boxes of $A^{*}$ in the frame. We make the following assumption about the spatial distribution of the proposals: the more the action proposals are on the same spatial location, the higher the likelihood that the action occurs in that location.
The use of action proposals for pseudo-annotations can be interpreted in two ways. First, it is a form of self-supervision~\cite{doersch2015unsupervised}, as we employ the action proposals to specify which action proposals to train on. Second, it is a form of outlier detection. If many proposals agree on the same location, we give a penalty to the proposals that are outside that location.

For each pixel $(x,y) \in F$, we denote the number of proposals from $A^{*}$ that contain $(x,y)$ as $C_{A^{*}}(x,y,F)$. We compute the pseudo-annotation as the center of mass over these counts:
\begin{equation}
p_{pa}(F) = \frac{1}{C_{A^{*}}(F)} \sum_{(x,y) \in F} C_{A^{*}}(x,y,F) \cdot (x,y),
\label{eq:pa}
\end{equation}
where $C_{A^{*}}(F)$ denotes the sum of the proposal counts over all pixels in $F$.

\textbf{Pseudo-annotations from frame centers.}
In~\cite{mettes2016spot,tseng2009quantifying}, it is noted that both actions and annotators have a bias towards the center of the video. We exploit this bias directly by adding a point-wise pseudo-annotation on the center of each frame of each video:
\begin{equation}
p_{fc}(F) = (F_W / 2, F_H / 2),
\end{equation}
where $F_W$ and $F_H$ denote the width and height of frame $F$.

\textbf{Pseudo-annotations from object proposals.}
The presence of objects is also correlated with the presence of actions, as observed in~\cite{jain2015objects2action,jain201515}.
Object proposals are computed here from EdgeBoxes~\cite{zitnick2014edge}, using the top 1,000 object proposals per frame.
Similar to the action proposal pseudo-annotation, we compute the number of proposals containing the pixel for each pixel in frame $F$. Let $C_{O}(x,y,F)$ denote the number of proposals containing pixel $(x,y)$, then the pseudo-annotation is given as:
\begin{equation}
p_{oa}(F) = \frac{1}{C_{O}(F)} \sum_{(x,y) \in F} C_{O}(x,y,F) \cdot (x,y).
\label{eq:oa}
\end{equation}
where $C_{O}(F)$ denotes the sum of the proposal count over all pixels in $F$. The difference with Equation~\ref{eq:pa} is in that we assume here that the foreground action is the most dominant object in the scene, as defined by the number of object proposals focusing on the action.

\subsection{Computing pseudo-annotation overlaps}
Each visual cue outputs an automatic box (person detection) or point (all others) annotation. Given an action proposal $A$, we compute the overlap with the box annotations using the spatial-temporal intersection-over-union score~\cite{jain2014action}. The intersection-over-union with a set of box annotations $B$ is computed as: $\frac{1}{|\Gamma|} \sum_{f \in \Gamma} \text{iou}(f_{B}, f_{A})$, where $\Gamma$ denotes the set of frames with at least one of $A$ and $B$ present.
For point-wise pseudo-annotation $P$, we compute the overlap using the function defined in~\cite{mettes2016spot}: $O(A, P, V)  =  M(A, P) - S(A, V)$. Here $M(A, P)$ states the overlap between action proposal $A$ and pseudo-annotations $P$ and is defined as: $M(A, P) = \frac{1}{|P|} \sum_{i=1}^{|P|}  \text{max} (0, 1 - \frac{||(P_{x_i},P_{y_i}) - \overbar{A_{P_i}} ||_2}{ \max_{(u,v) \in e(A_{P_i})} ||( (u,v) - \overbar{A_{P_i}}) ||_2})$, where $\overbar{A_{P_i}}$ denotes the center of the box of proposal $A$ in frame $P_i$. In turn, $S(A, V)$ is a size regularization on the action proposal itself: $S(A, V) = \big( \frac{ \sum_{i=f}^m |A_{i}| }{\sum_{j=1}^N |V_j|} \big) ^{2}$, where proposal $A$ runs from frame $f$ to frame $m$, $|\cdot|$ denotes the area of a box, and $V$ denotes the whole video. Intuitively, the overlap measure for point annotations aims to promote proposals with box centers close to the points while penalizing proposals of large size compared to the whole video.

\subsection{Correlation metric for pseudo-annotations}
For a video $v$, let $\{\textbf{S}_v^{(i)}\}_{i=1}^{|P|}$ denote the overlap scores over all pseudo-annotations $P$ and let $\textbf{S}_v^{(i)} \in \mathbb{R}^{|A_{v}|}$ denote the overlap scores for the $|A_{v}|$ action proposals of the $i^{th}$ pseudo-annotation in the video. Since no supervision within the videos is provided, it is \emph{a priori} unknown how pseudo-annotations from different cues should be used and to what extent.
Given the integral importance of people in detecting and localizing actions~\cite{yuCVPR2015fap}, we propose a correlation metric for pseudo-annotations using the person detection as an anchor for the correlation.

Let $\textbf{H}_v$ denote the overlap scores of the action proposals in video $v$ given by the person detection. Then we compute the statistical correlation between the pseudo-annotation of the $i^{th}$ cue and the pseudo-annotations from the person detection over all $N_t$ training videos:
\begin{equation}
\eta(P^{(i)}) = \frac{1}{N_t} \sum_{v=1}^{N_t} \frac{\text{cov}(\textbf{S}_v^{(i)}, \textbf{H}_v)}{\sigma(\textbf{S}_v^{(i)}) \cdot \sigma(\textbf{H}_v)}.
\label{eq:corr}
\end{equation}
The covariance and standard deviations in Eq.~\ref{eq:corr} are computed over the pseudo-annotation overlap scores of all action proposals in video $v$. Intuitively, Eq.~\ref{eq:corr} assigns a high score to pseudo-annotations that assign similar overlaps scores to the person detection; the more a pseudo-annotation agrees with the ranking of action proposals, the higher the correlation score. In turn, we can fuse the overlap scores of the pseudo-annotations as:
\begin{equation}
\textbf{S}_v^{\text{fused}} = \sum_{P^{(i)} \in P} \eta(P^{(i)}) \cdot [\![ \eta(P^{(i)}) \geq t ]\!] \cdot \textbf{S}_v^{(i)},
\end{equation}
where $t$ is a threshold to remove pseudo-annotations with overlap scores too dissimilar to the person detection.
Note that the person detection itself is also in the set $P$. In accordance with Eq.~\ref{eq:corr}, the person detection yields a correlation score of 1.

The correlation metric for pseudo-annotations provides a way of measuring the quality of pseudo-annotations without the need for manual box or point annotations, nor the need for examining test performance to combine and select pseudo-annotations.
By using a single pseudo-annotation per frame for each cue, we assume a single dominant action in each video. This assumption holds throughout our experiments. To handle videos with multiple actions and objects we can extend our approach with a density estimation over the pixel-wise weight of each cue to estimate multiple pseudo-annotations in frames.

\section{Experimental setup}

\subsection{Datasets}
\textbf{UCF Sports.} The UCF Sports dataset consists of 150 videos from sport broadcasts covering 10 action categories~\cite{RodriguezCVPR2008}, including \emph{Diving}, \emph{Riding a Horse}, and \emph{Skateboarding}. We employ the train and test data split as suggested in~\cite{lan2011discriminative}.
\\
\textbf{UCF 101.} The UCF 101 dataset has 101 actions categories \cite{soomro2012ucf101} where 24 categories have spatio-temporal action localization annotations. This subset has 3,204 videos, where each video contains a single action category, but might contain multiple instances of the same action. We use the first split of the train and test sets as suggested in~\cite{soomro2012ucf101}.
\\
\textbf{Hollywood2Tubes.} The Hollywood2Tubes dataset consists of 1,707 videos with ground truth point (training videos) and box (test videos) annotations~\cite{mettes2016spot}. The dataset contains the actions from the Hollywood2 dataset~\cite{marszalek09}, including \emph{Getting out of a car}, \emph{Hugging}, and \emph{Fighting}. We use the the train and test data split as suggested in~\cite{marszalek09}.
\\
\textbf{A2D.} The A2D dataset contains 3,782 videos of actions performed both by human actors and other actors, such as dogs, cars, and babies~\cite{xu2015can}. For a limited number of video frames, box annotations are provided. We use the train and test split as suggested in~\cite{xu2015can}.

We stress that throughout our experiments, we do not employ any manual point or box annotations for our approach.

\subsection{Implementation details}

\textbf{Proposals.} Following~\cite{mettes2016spot}, we employ the unsupervised action proposals from~\cite{vangemert2015apt}. We note that \cite{vangemert2015apt} only rely on dense trajectories for creating the proposals and do not use the cues that we employ for the pseudo-annotations.
\\
\textbf{Proposal representations.} On all datasets, we represent each action proposal with a Fisher Vector~\cite{sanchez2013image} with 128 clusters over the improved dense trajectories~\cite{wang13} within the proposal. This results in a 54,656-dimensional representation per proposal.
\\
\textbf{Training.} We train the Multiple Instance Learning algorithm for 5 iterations for all evaluations. Following~\cite{cinbis2014multi}, we split the training videos into multiple folds during training for the classifier and proposal selection steps. We set the regularization parameter $\lambda$ in the max-margin optimization to 10 in all experiments.
\\
\textbf{Evaluation.} During testing we apply the classifier of an action to all proposals of a test video and keep the proposal with the highest classifier score~\cite{jain2014action,vangemert2015apt}. To evaluate the action localization performance, we compute the intersection-over-union in space and time between the top proposal and a ground truth tube as defined in~\cite{jain2014action}. Only  proposals whose overlaps with ground truths exceed the threshold are considered correct.

\section{Experimental results}

\subsection{Evaluating the pseudo-annotations}

In the first experiment, we evaluate each pseudo-annotation individually for action localization on UCF Sports and UCF-101 with the mean Average Precision score. We compare the pseudo-annotations to two baselines. The first baseline uses full box supervision during training (light gray area). This baseline serves as a supervision upper bound. The second baseline uses the video labels with standard Multiple Instance Learning (dark gray area). This baseline serves as the supervision lower bound. Note that all approaches use the same features and classifier settings.


\begin{figure}[t]
\centering
\includegraphics[width=0.475\textwidth]{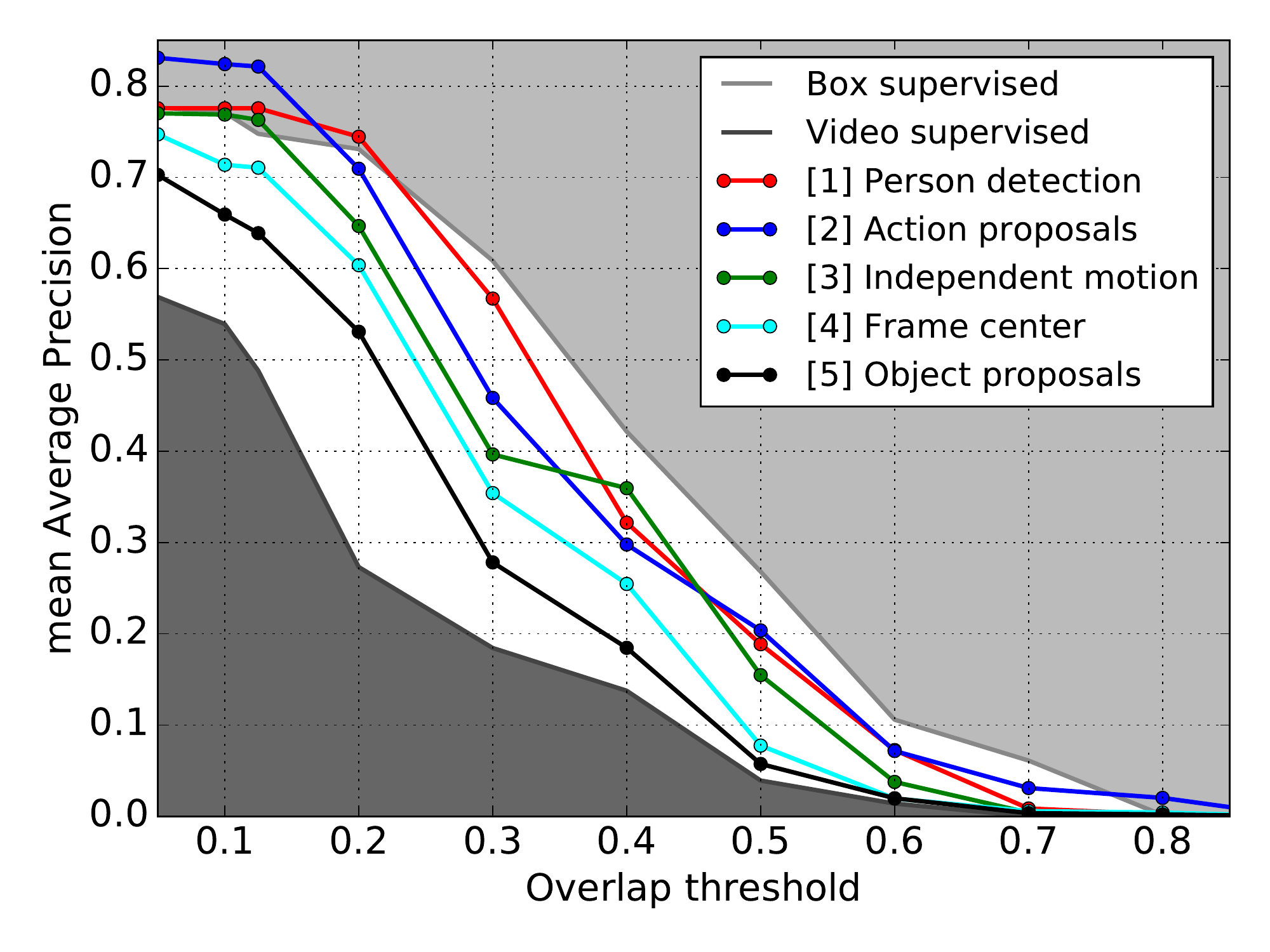}
\includegraphics[width=0.475\textwidth]{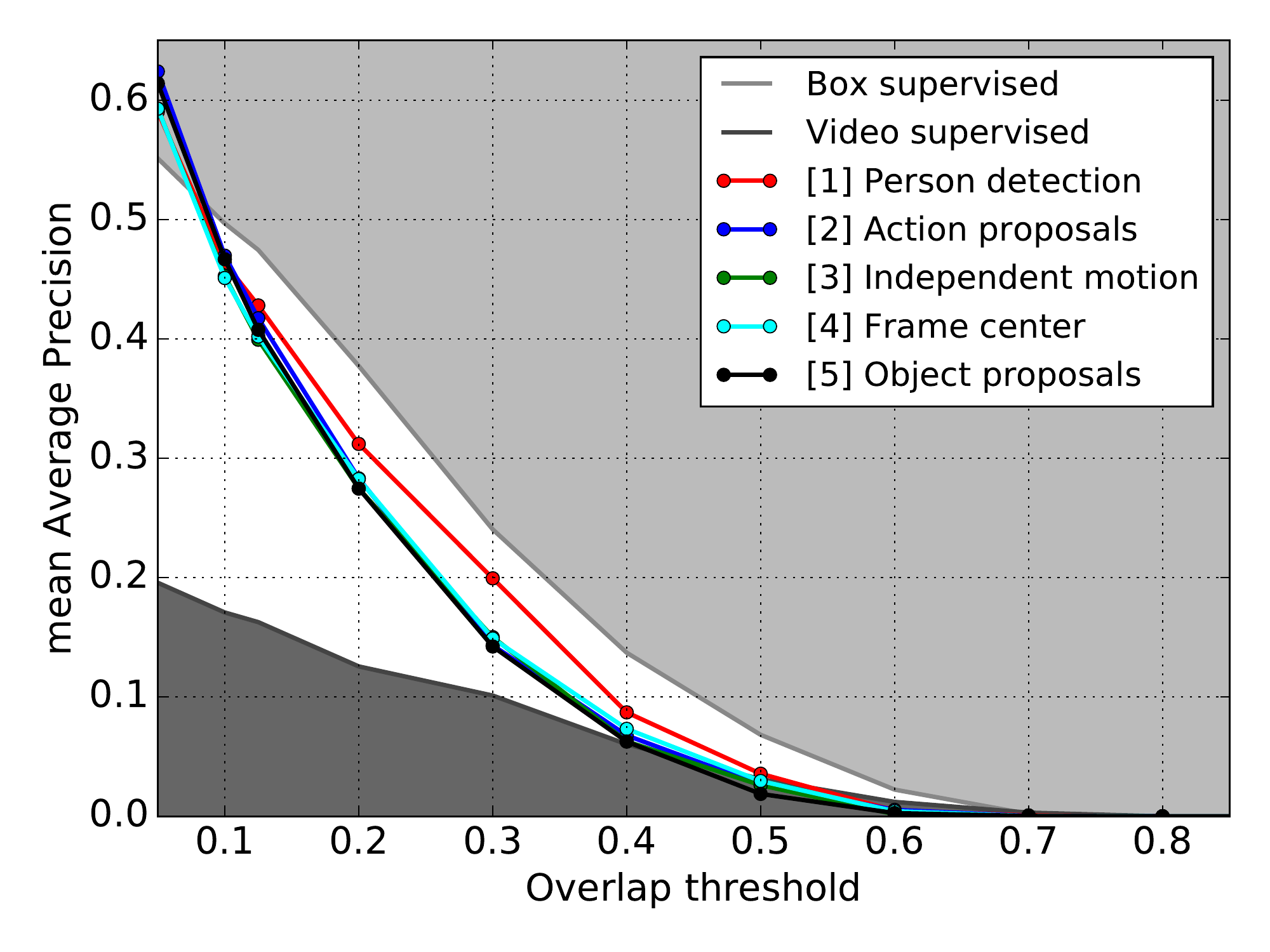}
\caption{\textbf{Pseudo-annotation action localization performance} on UCF Sports (left) and UCF-101 (right), compared to the supervision upper and lower bounds.}
\label{fig:exp1}
\end{figure}

The scores across all overlap thresholds are shown in Figure~\ref{fig:exp1}. On UCF Sports, we observe that each pseudo-annotation performs better than only using the video label, which means that pseudo-annotations provide meaningful information about the location of actions in videos. Furthermore, the person detection performs best, followed by action proposals and independent motion. At low overlap thresholds, these approaches even outperform full supervision. This is surprising, since no human intervention is provided. At higher thresholds, full supervision is still better, while all approaches break down at the highest thresholds.

\begin{figure}[t]
\centering
\subfigure[Correct pseudo-annotations.]{\includegraphics[width=0.24\textwidth]{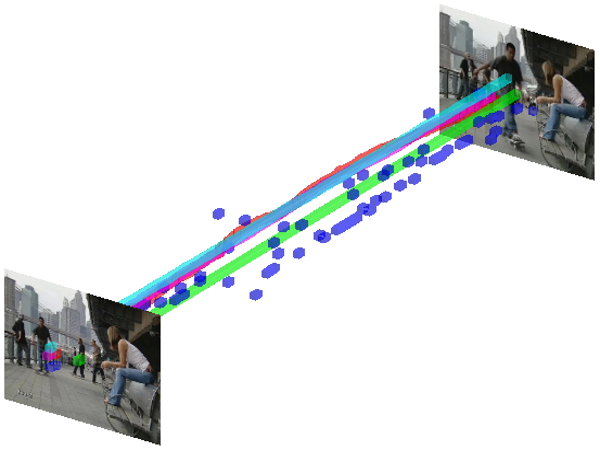} \includegraphics[width=0.24\textwidth]{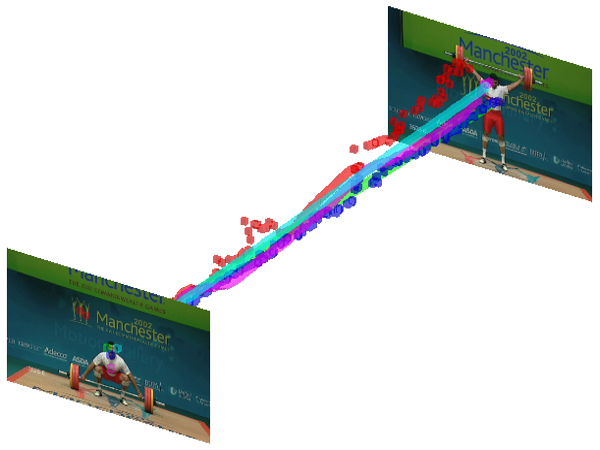}\label{fig:qual-a}}
\hspace{0.15cm}
\subfigure[Incorrect pseudo-annotations.]{\includegraphics[width=0.24\textwidth]{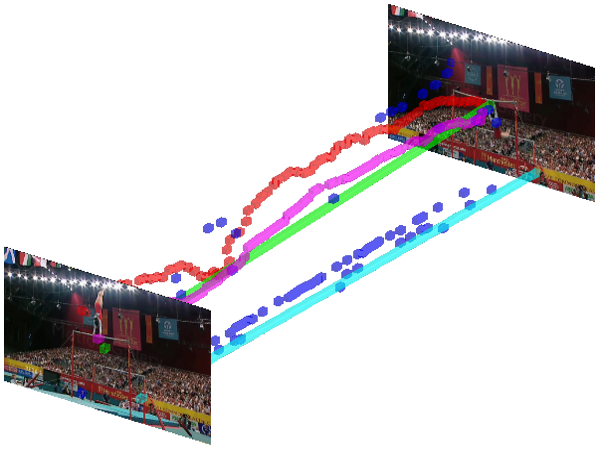} \includegraphics[width=0.24\textwidth]{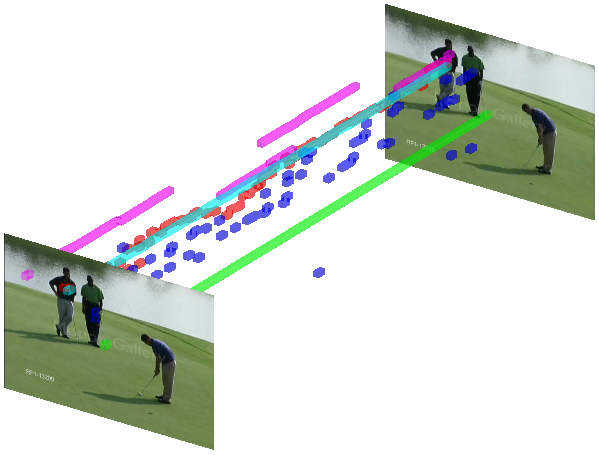}\label{fig:qual-b}}
\caption{\textbf{Qualitative examples of pseudo-annotations} in non-trivial action videos. In (a), the pseudo-annotations successfully follow the centrally oriented main action. In (b), some pseudo-annotations are distracted either by complex background (\emph{Swinging on a bar}, left), or due to lack of primary action and the presence of other people (\emph{Golf swinging}, right).}
\label{fig:qual}
\end{figure}

On UCF-101, the pseudo-annotations similarly all outperform the approach using the video label only. The difference between the approaches is smaller since the dataset is larger, making it more robust against accidental hits and misses of the pseudo-annotations. The person detection pseudo-annotation performs best, followed by using frame centers and action proposals.
To highlight the effect and limitations of the pseudo-annotations, we show qualitative results in Figure~\ref{fig:qual}.

\subsection{Combining pseudo-annotations}
\begin{figure}[t]
\centering
\subfigure[Correlations (UCF Sports).]{\includegraphics[width=0.32\textwidth,height=0.17\textheight]{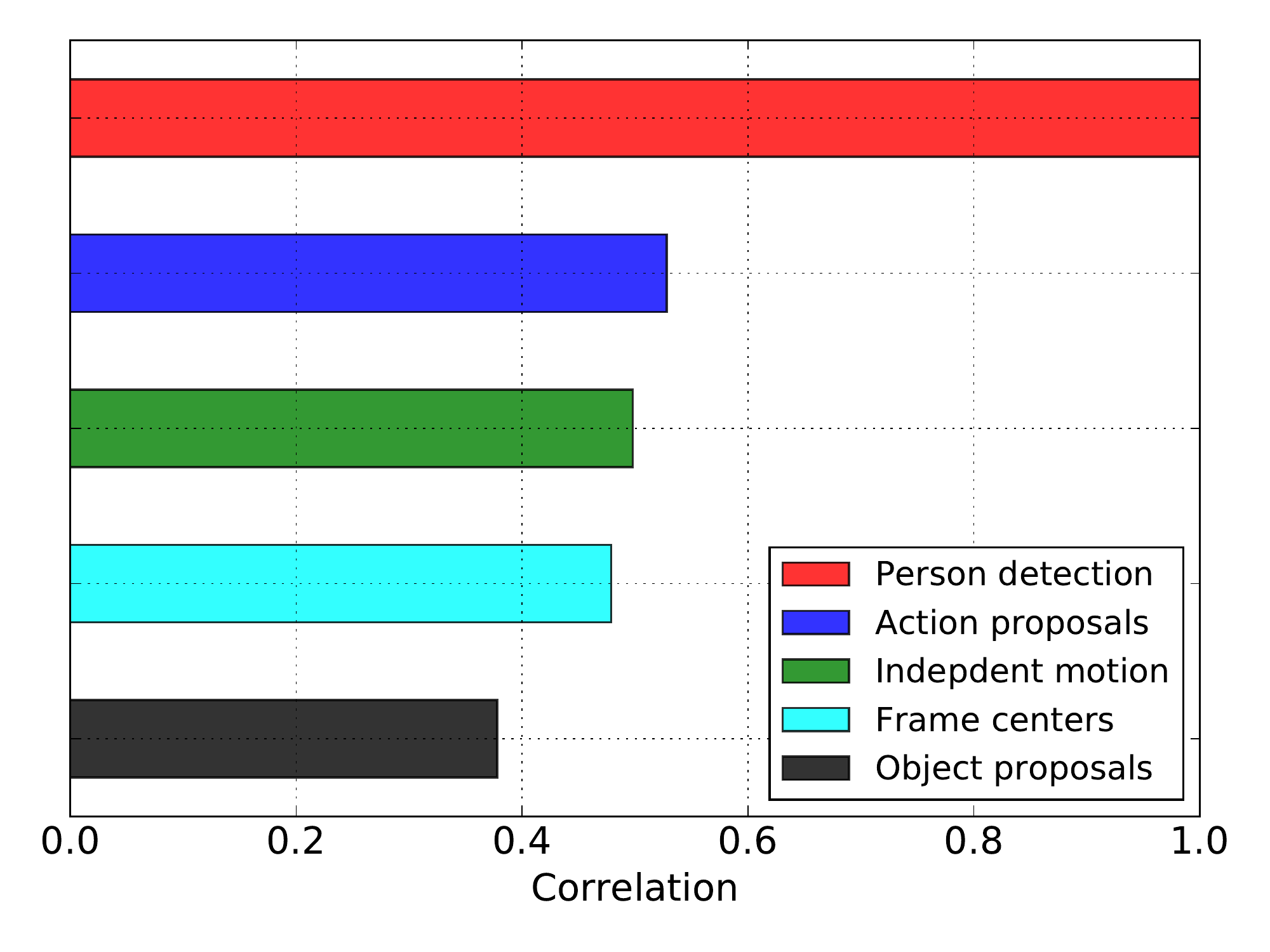}\label{fig:exp2-a}}
\subfigure[UCF Sports.]{\includegraphics[width=0.32\textwidth,height=0.17\textheight]{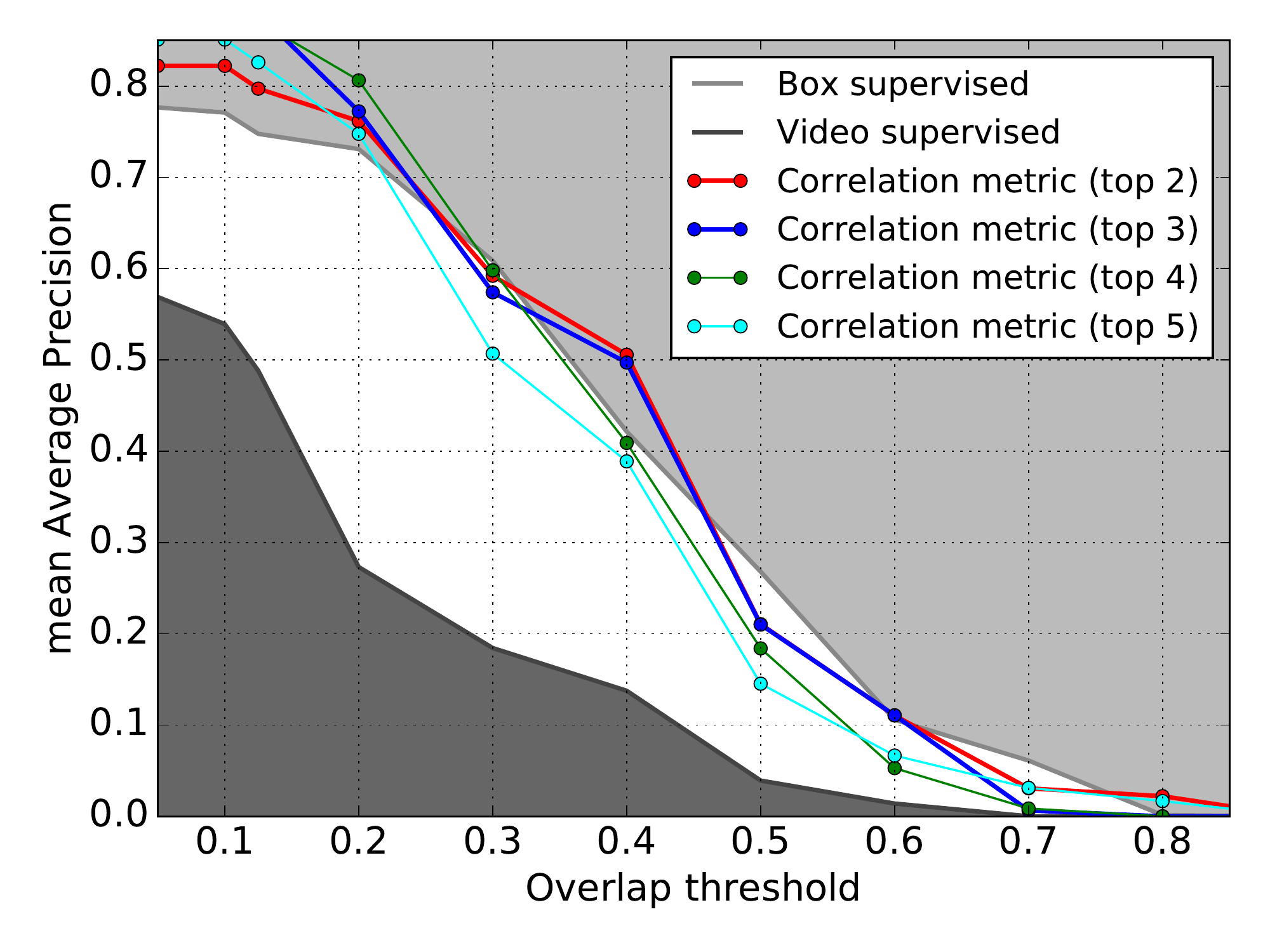}\label{fig:exp2-b}}
\subfigure[UCF-101.]{\includegraphics[width=0.32\textwidth,height=0.17\textheight]{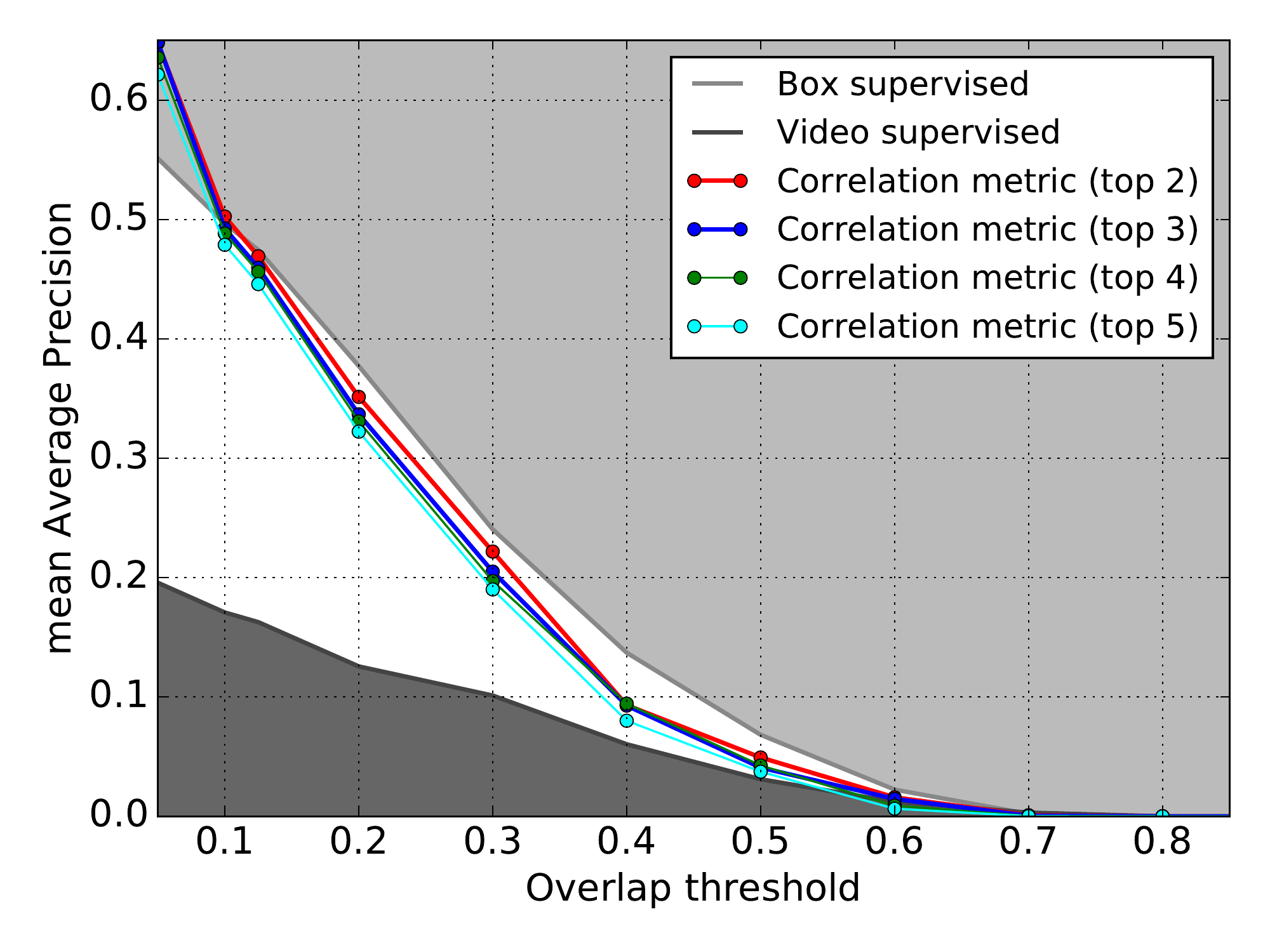}\label{fig:exp2-c}}
\caption{\textbf{Correlation-based combination of pseudo-annotations.} On UCF Sports and UCF-101, automatically combining the top two or three correlated pseudo-annotations yields the best results, even comparable to full box supervision.}
\label{fig:exp2}
\end{figure}

In the second experiment, we evaluate the correlation metric for pseudo-annotations.
In Figure~\ref{fig:exp2-a}, we show the correlation scores on UCF Sports.
The scores show that person detection, action proposals, and independent motion pseudo-annotations are most relevant, while frame centers and object proposals are less relevant. The discovered order is in line with the order of performance from the first experiment.
This means that the correlation metric provides a reliable way of measuring the quality of pseudo-annotations, while automatic selection is possible by using the ones with highest average correlation.

We provide the localization performance in Figures~\ref{fig:exp2-b} and~\ref{fig:exp2-c}. For UCF Sports, when using the correlation metric with both the top-two and top-three pseudo-annotations yields results comparable but not identical to full box supervision. When using more pseudo-annotations, the performance at higher overlap thresholds degrades, indicating that not all pseudo-annotations should be used in the combination.
On UCF-101 (Figure~\ref{fig:exp2-c}), using correlation metric with the top pseudo-annotations also yields results comparable or close to full box supervision. Here, the combination using person detection and frame centers (the second highest correlation pseudo-annotation) performs best. Incorporating more pseudo-annotations slightly degrades the performance.

We conclude from this experiment that a correlation metric from the top pseudo-annotations provides a reliable way to merge different visual cues for action localization. On both datasets, the metric with the top 2/3 pseudo-annotations outperform person detections only, while performing comparable to full box supervision.

\textbf{Non-human action localization.}
The datasets typically used in action localization are human-centric~\cite{mettes2016spot,RodriguezCVPR2008,soomro2012ucf101}.
Here, we evaluate how well our pseudo-annotations generalize to actions performed by non-human actors using the A2D dataset~\cite{xu2015can,xu2016actor}.
These actors include babies, balls, birds, cars, cats, and dogs.
Since this dataset does not have action tube annotations, our approach can not be directly evaluated.
Individual box annotations for a set of frames per video are provided instead.
Therefore, we investigate whether pseudo-annotations are capable of "pointing at" actions performed both by human and non-human actors.
We evaluate how the overlap between proposals and pseudo-annotations relates to the overlap between proposals and ground truth boxes.

Over all actors, we find that the Pearson correlation score is 0.29; a high score for the pseudo-annotations correlates with a high score in action overlap, which strengthens the notion of pseudo-annotations for action localization.
We also find that the Pearson correlation score is positive for all actor types individually.
We do note that the score is higher for the person actor than the other types, indicating a bias towards persons as actors in our pseudo-annotations.
Interestingly, when excluding the person detection as pseudo-annotation, person remains the most postively correlated actor type.
We conclude that our pseudo-annotations are not restricted to the person as actor type and handle other actor types as well.
The person as actor type does have closest relations to the pseudo-annotations, although this is not solely due to the use of the person detection itself.

\begin{table}[t]
\centering
\scalebox{0.8}{
\centering
\begin{tabular}{l | r r r r r r | r r r r r r}
\toprule
 & \multicolumn{6}{c}{UCF Sports} & \multicolumn{6}{c}{UCF 101}\\
 & \textbf{0.1} & \textbf{0.2} & \textbf{0.3} & \textbf{0.4} & \textbf{0.5} & \textbf{0.6} & \textbf{0.1} & \textbf{0.2} & \textbf{0.3} & \textbf{0.4} & \textbf{0.5} & \textbf{0.6}\\
\midrule
Full box annotations & 77.1 & 73.1 & 60.9 & 42.2 & 26.8 & 10.6 & 46.1 & 34.5 & 24.4 & 13.9 & 7.8 & 1.9\\
\midrule
Pseudo-annotations & 88.0 & 77.2 & 57.4 & 49.7 & 21.0 & 11.1 & 50.3 & 35.1 & 22.2 & 9.3 & 4.9 & 1.6\\
Pseudo-annotations $++$ & 87.7 & 81.7 & 64.4 & 54.5 & 37.8 & 17.5 & 49.8 & 37.4 & 25.8 & 13.7 & 6.2 & 1.3\\
\midrule
\midrule
Full box annotations $++$ & 86.7 & 86.7 & 74.0 & 61.2 & 42.3 & 23.1 & 50.6 & 40.8 & 28.8 & 17.5 & 8.3 & 2.4\\
\bottomrule
\end{tabular}
}
\caption{\textbf{Localization performance (\%) with pseudo-annotations during testing} (++). Using pseudo-annotations during testing increases the performance across all overlap thresholds and datasets, even outperforming full box supervision. Pseudo-annotations can also be employed to improve models trained with full box supervision.}
\label{tab:exp3}
\end{table}

\subsection{Pseudo-annotations at test time}
Since pseudo-annotations are automatically generated for videos, their use is not restricted to training videos only. In the third experiment, we employ pseudo-annotations during testing to help select the best proposal per video. We do this by combining the classifier score with the overlap scores from the pseudo-annotations. We employ the correlation metric with the top pseudo-annotations for this experiment.

Results on UCF Sports and UCF-101 are shown in Table~\ref{tab:exp3}. On both datasets, we observe a jump in performance when adding pseudo-annotations during testing, even outperforming the full box supervision results. This performance shows the effectiveness of the pseudo-annotations for action localization. We also evaluate the effect of using pseudo-annotations during testing with a model trained on full box supervision, which yields a similar increase in performance. We conclude from this experiment that pseudo-annotations during testing improves any model trained on unsupervised proposals. With only the video labels as manual annotations, we even outperform the standard full box supervision setup.

\subsection{Comparison to state-of-the-art}
We compare our results on three action localization datasets to the current state-of-the-art. In Table~\ref{tab:sota}, we show the performance of the methods, ordered by their level of supervision. To maximize the number of comparisons, we show the results at a threshold of 0.2.

On all datasets, we perform comparable to approaches that rely on expensive manual box or point annotation during training and unsupervised proposals during testing. This result is encouraging, as it states that video labels and automatic pseudo-annotations can provide enough information for localization. We improve over approaches using only video labels~\cite{cinbis2014multi} or zero-shot information~\cite{jain2015objects2action}, resulting in state-of-the-art performance on the Hollywood2Tubes dataset. We also compare against the approaches of Weinzaepfel \etal~\cite{weinzaepfelICCV2015learningToTrack} and Saha \etal~\cite{saha2016deep}. On UCF Sports, we achieve comparable AUC scores. On UCF-101, these approaches report higher scores. While effective, these approaches require full box supervision both for making proposals and training action classifiers. They can therefore not generalize to weaker forms of supervision.

\begin{table}[t]
\centering
\scalebox{0.79}{
\begin{tabular}{l l l r r r}
\toprule
 & & & \multicolumn{1}{c}{UCF Sports} & \multicolumn{1}{c}{UCF 101} & \multicolumn{1}{c}{H2T}\\
\textbf{Method} & \textbf{Proposal annotations} & \textbf{Classifier annotations} & \textbf{AUC} & \textbf{mAP} & \textbf{mAP}\\
\midrule
Wang \etal~\cite{wang2014video} & n.a. & video-label + joints & 47.0 & - & n.a.\\
Saha \etal~\cite{saha2016deep} & video-label + boxes & video-label + boxes & - & 66.8 & n.a.\\
Weinzaepfel \etal~\cite{weinzaepfelICCV2015learningToTrack} & video-label + boxes & video-label + boxes & 55.9 & 46.8 & n.a.\\
\midrule
Jain \etal~\cite{jain2014action} & none & video-label + boxes & 52.0 & - & n.a.\\
van Gemert \etal~\cite{vangemert2015apt} & none & video-label + boxes & 54.6 & 34.5 & n.a.\\
Mettes \etal~\cite{mettes2016spot} & none & video-label + points & 54.5 & 34.8 & 14.3\\
Cinbis \etal~\cite{cinbis2014multi},\cite{mettes2016spot} & none & video-label & 27.8 & 13.6 & 0.9\\
Jain \etal~\cite{jain2015objects2action} & none & zero-shot & 23.2 & - & -\\
\midrule
This paper & none & video-label & 53.3 & 35.1 & 13.6\\
This paper $++$ & none & video-label & 55.6 & 37.4 & 17.2\\
\bottomrule
\end{tabular}
}
\caption{\textbf{Localization results (\%)} at an overlap of 0.2. A dash (-) states that results are not provided, while n.a. states that the approach can not be applied due to the dataset's lack of required annotations. The sign ($++$) denotes the use of pseudo-annotations during testing.
We achieve results comparable to approaches that train on unsupervised proposals and box annotations, while outperforming approaches using video labels or zero-shot information considerably.}
\label{tab:sota}
\end{table}


\section{Conclusions}
In this work, we introduce pseudo-annotations for localizing actions in videos.
We investigate pseudo-annotations from person detection, independent motions, action proposals, center biases, and object proposals.
Using a correlation metric for pseudo-annotations, we reach results comparable or better to using full box supervision with the same settings, while outperforming other weakly-supervised approaches.
As our approach relies on action class labels as the only manual annotations, it enables action localization on any action classification dataset, such as Sports 1M~\cite{karpathy2014large}, ActivityNet~\cite{caba2015activitynet}, and EventNet~\cite{ye2015eventnet}.

\section*{Acknowledgements}
\vspace{-0.25cm}
This research is supported by the STW STORY project.

\bibliography{egbib}

\end{document}